\documentclass{article}

\usepackage{microtype}
\usepackage{graphicx}
\usepackage{subfigure}
\usepackage{booktabs}
\usepackage{hyperref}

\usepackage[accepted]{icml2026}

\usepackage{amsmath}
\usepackage{amssymb}
\usepackage{mathtools}
\usepackage{amsthm}
\usepackage{enumitem}
\DeclareMathOperator*{\argmax}{argmax}
\DeclareMathOperator*{\argmin}{argmin}

\usepackage[capitalize,noabbrev]{cleveref}

\theoremstyle{plain}

\theoremstyle{definition}

\theoremstyle{remark}

\usepackage[textsize=tiny]{todonotes}

\begin{document}

\twocolumn[
\icmltitle{A Definition of Good Explanations and the Challenges Explaining LLM Outputs}



\icmlsetsymbol{equal}{*}

\begin{icmlauthorlist}
\icmlauthor{Louis Mahon}{yyy}
\icmlauthor{Elliot Ford}{yyy}
\icmlauthor{Callum Hackett}{yyy}
\end{icmlauthorlist}

\icmlaffiliation{yyy}{UnlikelyAI, London, UK}
\icmlcorrespondingauthor{Louis Mahon}{louis@unlikely.ai}

\icmlkeywords{Machine Learning, ICML}

\vskip 0.3in
]

\printAffiliationsAndNotice{} 

\begin{abstract}
How to define a good explanation is a long-standing philosophical debate which has found recent renewed interest in the context of AI outputs. Explainability is crucial for AI adoption in many contexts, but in order to produce good explanations of AI systems, we must first have an understanding of what good explanations are. In this paper we propose a definition inspired by the notion of counterfactual explanations, however we argue that one must also take into account the interlocutor's prior beliefs in each fact that could be offered in an explanation. 
We explore the ramifications of this definition for AI explainability and, in particular, why LLM outputs are difficult to produce good explanations for.
\end{abstract}

\section{Introduction: Counterfactual Explanations}
The concept of an explanation is widely used but surprisingly hard to pin down. Various approaches have been made to defining explanations, and one, which we take as the starting point for this paper, is the notion of counterfactuals.

Going back to \citeauthor{hume1748enquiry} (\citeyear{hume1748enquiry}), counterfactuals have featured prominently in philosophical contexts, mostly as a way to define the notion of causality \cite{pearl2009causality}, especially in the sciences \cite{lewis1973causation, woodward2005making}: a counterfactual cause of a given effect is something which, had it not happened, would have meant the effect did not happen either. Counterfactual explanations extend this idea to the more general concept of explanations, which may sometimes include non-causal elements \cite{reutlinger2016there}. ML explanations lend themselves well to a counterfactual account. For example, they are described in \citet{leofante2023counterfactual} as ``showing how an input could be changed to produce a different decision'', or similarly in \cite{guidotti2024cfes, wachter2017counterfactual, doshi2017towards}.

Although counterfactuals are the basis for the definition we propose, we will argue that a good explanation also needs to take account of the interlocutor's prior beliefs. This will lead us to a probabilistic definition of a good explanation that can be expressed mathematically. Section \ref{sec:proposed-definition} motivates and proposes our definition, considers implications for the general concept of an explanation, and addresses some potential problems. Section \ref{sec:explanations-in-LLMs} discusses the principled difficulties of explaining LLM outputs in the context of this definition.

\section{Proposed Definition} \label{sec:proposed-definition}
As a starting point, we can consider the following basic definition of what it means to explain something:

\begin{description} \label{def:h1}
  \item[H1] An explanation of an observation $O$ is a fact which, if it were not true and all else was equal, would mean $O$ would not have been observed.
\end{description}

For example, Bob comes into the office with his hair wet. If someone asks ``why is his hair wet?'', we can answer ``because it is raining'', provided that if it had not been raining, his hair would not be wet. In this definition, we do not appeal to any specific interpretation of the logical relation between the observation and the explanatory counterfactual (such as causation). We instead approach the issue more pragmatically, allowing such facts and interpretations as are contextually appropriate.

However, an issue immediately arises: there are many facts which satisfy H1 but which seem as though they are not explanatory: Bob did not towel his hair right before entering the room; there was not a drone flying above him as he walked that kept him dry; there was not a very strong and hot wind on the stairs which dried him off, etc. On the face of it, these facts function less like explanations and more like preconditions for the possibility of the outcome.

It may seem like we can avoid these examples by restricting the counterfactuals to positive facts, i.e. those that do not negate some state of affairs. However, we do have to allow for negative facts in some explanations: why is  Alice hungry? because she did not eat anything yesterday; why was the bicycle stolen? because it wasn`t locked. Also, there are positive facts which similarly lack explanatory value: why is Bob's hair wet? because his hair is a substance that water adheres to, 
or because getting rained on makes you wet. 

Rather than negation being the reason these examples are defective as explanations, we claim that the problem is that they would already have been believed with high probability by the interlocutor.
This motivates an updated definition: 

\begin{description} \label{def:h2}
  \item[H2] An explanation of an observation $O$ is a fact which, if it were not true and all else was equal, would mean $O$ would not have been observed, and on which we believe the interlocutor placed low probability.
\end{description}

Under certain reasonable assumptions about the interlocutor`s beliefs, H2 rules out all the above examples of bad explanations but none of the examples of good ones. 

Given that they are asking for an explanation, we can assume that $O$ was sufficiently surprising for them to desire updates to their priors on the relevant facts. The task of producing a good explanation, then, comes down to identifying those facts on which the prior was especially low, and communicating with the interlocutor to increase those probabilities. This approach to explanations essentially obeys the cooperative principle in \citet{grice1975logic}.

Our injunction to select facts with low prior belief by the interlocutor aligns, on a high level, with the discussion of explanations by \citet{miller2019explanation}, who invokes the concept of contrastive explanations (originally from \citet{lipton1990contrastive}). Contrastive explanations aim to explain why a given thing happened instead of something else that was expected and is referred to as the foil. In our framework, the facts with high prior probability analogise roughly to those that would have produced the foil.

\subsection{Formalisation} \label{subsec:formalisation}
Let $\mathcal{F}$ be some predefined set of possible facts under consideration. Let $CE(f,O)$ be an indicator that fact $f$ is a counterfactual explanation of $O$, by H1, and define $C(\mathcal{F}, O) = \{f \in \mathcal{F} | CE(f,O)\}$. Then a good explanation of observation~$O$ to interlocutor $a$, with respect to $\mathcal{F}$, written $X(a,O,\mathcal{F})$, is given by
\begin{equation} \label{eq:h2}
    X(a,O,\mathcal{F}) = \argmin_{f \in C(\mathcal{F}, O)} p(a,f)\,,
\end{equation}
where $p(a,f)$ is the probability that $a$ places on $f$.\footnote{Given we are allowing negative facts, this communication may involve a negated statement, and the effect on the interlocutor of raising their belief in a negative fact could also be modelled as lowering their belief in the corresponding positive fact.}

Let $p(a,O)$ be $a$'s predicted probability that they were going to observe $O$, and $\bar{p}(a,\mathcal{S})$ be the joint probability of all facts in set $\mathcal{S}$ being true. Then $\bar{p}(a,C(\mathcal{F}, O)) \leq p(a,O)$. The reason it is a lower bound, rather than an equation, is that there could be other routes by which $a$ could have expected to observe $O$. 

We have assumed the facts to be semantically independent, and H1/2 include the condition that we inform the interlocutor of a given fact but assume all else is equal. Arguably, this implies a certain form of statistical independence, in particular, that we can adjust their belief in a given fact without affecting their beliefs in the others, which in turn suggests the following independence statement:
\begin{gather} 
    \forall f \in C(\mathcal{F}, O) \nonumber \\
    \bar{p}(a, C(\mathcal{F}, O) \setminus \{f\})p(a,f) = p(a, C(\mathcal{F}, O))\,, \label{eq:varwise-indep}
\end{gather}
and repeated application of \eqref{eq:varwise-indep} implies full independence.

We need not believe independence literally holds between the probability $a$ places on each fact.
However, as a simple extension of the concept of independent facts, and the criterion in H1/2 of assuming all else equal, we \emph{model} their beliefs as if $\bar{p}(a, C(\mathcal{F}, O) \setminus \{f\})$ does not change when we change $p(a,f)$. The alternative requires considering possible effects of stating one fact on the interlocutor`s degree of belief in all other facts, and this makes explanations prohibitively difficult to reason about.

We can then formulate \eqref{eq:h2} alternatively in terms of the remaining facts:
\begin{gather} 
    \forall f \in C(\mathcal{F}, O), \bar{p}(a, C(\mathcal{F}, O) \setminus \{f\}) = \frac{\bar{p}(a, C(\mathcal{F}, O))}{p(a,f)} \nonumber \\
 \Rightarrow \nonumber \\
    \argmin_{f \in C(\mathcal{F}, O)} p(a,f) = \argmax_{f \in C(\mathcal{F}, O)} \bar{p}(C(\mathcal{F}, O) \setminus \{f\}, a)\,. \label{eq:argmin-argmax-equivalence}
\end{gather}
So the injunction of H2 to use as explanation the lowest prior element of $C(\mathcal{F},O)$ is equivalent to maximising the post-communication value of $\bar{p}(a, C(\mathcal{F}, O) \setminus \{f\})$.

If we further assume that, after $a$ being told $f$, their belief that $f$ is not true shrinks by a fixed fraction, then we can show that H2 is equivalent to maximising a lower bound on their (post-communication) belief that they were going to observe~$O$. In the simplest case, where their belief $f$ is not true shrinks to zero and $p(a,f)$ becomes~1, $\bar{p}(a, C(\mathcal{F}, O) \setminus \{f\})$ is this lower bound, and so \eqref{eq:argmin-argmax-equivalence} shows that the bound is maximised. 


More generally, let $\epsilon$ denote the fraction of uncertainty that remains, so that, after $a$ being told $f$, $p(a,f)$ becomes $1 - \epsilon(1-p(a,f))$. Then, the effect of being told $f$ is to multiply the lower bound on the interlocutor`s (post-communication) probability that they were going to observe~$O$ by $\frac{1 - \epsilon(1-p(a,f))}{p(a,f)}$, c.f. multiplying by $\frac{1}{p(a,f)}$ if assuming testimony yields certainty. Either way, the factor is a monotonically decreasing function of $p(a,f)$, meaning H2 is equivalent to maximising the lower bound.

\subsection{Subjectivity of Explanations} \label{subsec:subjectivity-of-explanations}
As people may differ in their priors over the relevant facts, H2 implies that a good explanation may be different for different interlocutors. If, for example, the person asking why Bob`s hair was wet was, at that time, looking out the window at the rain
then ``it is raining'' would not be a good explanation. We would instead search for another fact that meets H2, such as ``he didn't have an umbrella''. The dependence of a good explanation on the context of the listener has been argued for by \citet{vanFraassen1980pragmatics} and others. Our framework articulates that the specific source of context-dependence is the prior beliefs of the interlocutor.

This does not mean, however, that we need explicit evidence for the interlocutor having a certain prior on each fact. Rather, we assume a default value for their priors, which can be updated in the presence of appropriate evidence, such as seeing them look out the window at rain. One simple source for obtaining default priors is to assume they are the same as your own, e.g. you do not, a priori, think it likely that someone has a drone flying above them while walking to work, so you assume the interlocutor's prior on this fact is also low. For more fleshed out accounts of modelling the beliefs of others, various theories have been posited, such as simulation theory \cite{gordon1986folk} or the principle of charity/humanity \cite{davidson1975thought,grandy1973reference}. Default values for the other's priors also relates to the concept of common ground \cite{stalnaker2002common}. 
Using H2 does not depend on a particular theory of folk psychology, only that we have \emph{some} means of obtaining a reasonable estimate for the interlocutor`s relevant priors. 

\subsection{Explanations with Multiple Facts} \label{subsec:explanations-with-multiple-facts}
Sometimes, there may be multiple facts satisfying H2, meaning that, if we just selected the argmin, as in \eqref{eq:h2}, the explanation would be incomplete in the sense that there would still be other facts that meet H2 but have not been conveyed. This then raises the question of how many facts are appropriate to include. The interlocutor`s predicted probability on $O$ is unlikely to reach 1 even after a good explanation, but including all elements of $C(\mathcal{F},O)$ would produce an overly long explanation. When allowing multiple facts, we have to include concision as a second criterion, alongside maximising their $p(a,O)$. 
One could attempt to model this by setting $k$ as the number of facts to include, and then replacing the argmax in \eqref{eq:h2} with topk. However, this would mean all explanations have to have the same number of facts, which does not seem correct. A better way to model the dual objectives is to set 
a threshold that we want $p(a,O)$ to reach, and then include the minimum number of facts to reach that threshold:\footnote{If this does not produce a unique subset, we could select the lowest $\prod_{g \in F'} p(a,g)$, or simply regard all as good explanations.}

\begin{equation} \label{eq:h2-max}
    X(a,O,\mathcal{F},p^*) = \argmin_{\{F' \subseteq C(\mathcal{F}, O) | \bar{p}(a, C(\mathcal{F}, O) \setminus F') \geq p^*\} } |F'|\,.
\end{equation}

As $p(a,C(\mathcal{F}, O))$ is a lower bound, this guarantees $p(a,O) \geq p^*$.\footnote{This is again assuming testimony yields certainty. Weakening that assumption, as in Section \ref{subsec:formalisation}, to be that the interlocutor moves a fixed fraction of the way towards certainty, we could modify \eqref{eq:h2-max} so that set over which the argmin is taken is over to include $\frac{1 - \epsilon(1-p(a,f))}{p(a,f)}$ terms, and so produce an equivalent result.}

\subsection{The Curious Case of XOR} \label{subsec:xor-example}
There is a suggestion, when considering multiple counterfactuals, that if any subset of them held, $O$ would not have been observed. However, this is not always true. For example, suppose two attempts were made to poison Charlie on the same night, one in his water and one in his food. He happens to not consume either, and so lives. The following are then facts which explain Charlie being alive:

\begin{description} \label{def:f_1_f_2}
  \item[$f_1$] Charlie did not drink his water.
  \item[$f_2$] Charlie did not eat his dinner.
\end{description}

Suppose also, however, that the substances of the poisons are such that, if both ingested together, a complex pharmacological interaction means they are less deadly than either alone (atropine and organophosphates could potentially be an example). Then, if $f_1$ and $f_2$ were false, Charlie would be sick but still alive. The logical structure here is XOR: Charlie is alive if and only if $f_1 \otimes f_2$. 

We still contend that a good explanation of why Charlie survived is ``he did not drink his water and he did not eat his dinner'', so H2 still applies. (It is still true that $\bar{p}(a, C(\mathcal{F}, O))  \leq p(a,O)$.) However, such explanations make no claims about what would have happened if he had drunk his water \emph{and} eaten his food. The definition of $C(\mathcal{F},O)$ is a set of facts such that, if all are true, then $O$ will be observed, and if exactly one is false and all else is equal, then it will not. This means it is important that we can decompose into a set of facts and reason counterfactually about each of them independently. Considering only a single fact $f' = f_1 \land f_2$, would not allow us to explain why Charlie survived. This will be relevant in Section \ref{sec:explanations-in-LLMs}.

\subsection{Identifying the Relevant Facts} \label{subsec:identifying-relevant-facts}
H2 also requires a set of facts under consideration, written $\mathcal{F}$ in \eqref{eq:h2}. In general, identifying $\mathcal{F}$ is a hard problem. The reasons why are similar to those that make the frame problem hard, namely that the set of facts that potentially could affect whether $O$ was observed is very large, and mostly composed of facts that humans do not even entertain. 

If H2 were to be put forward as a general answer to the question of what makes a good explanation, an account would need to be given of how $\mathcal{F}$ could be identified.\footnote{We expect progress could be made by saying that we generate new facts on the fly, rather than having all candidate facts available and then select the one with the lowest prior. That is, try to come up with something that satisfies H2, if successful, use it as the explanation, if not, try to come up with something else etc. Thus, \eqref{eq:h2} would use a generator instead of a static set $\mathcal{F}$.} However, in the case of explanations for automated systems, we can sidestep this problem. 
A user provides certain information to the system and receives an output, then wants to ask why that output was produced. We assume that the answer must relate to either the input or to the system. For example, we can exclude, from $\mathcal{F}$, general facts about the world, computers or mathematics. The inclusion of facts about the model differs from most works on counterfactual explanations in ML, which search only for changes to the input.

As an example of a fact about the system, suppose a bank customer applies for a loan and provides information about their age and income, and is rejected by the bank`s automated system. An explanation could be that the system rejects all applicants with income below a certain threshold (which the customer`s income was below). 
Such a fact could itself admit further explanations, which might, in the case of an ML model, invoke facts about training data or hyperparameters. However, characterising those sorts of explanations is not our focus here, and if one was going to apply H2, they would need a different way to constrain the set of facts that are to be considered as counterfactuals.

An example where the explanation includes facts about the input might be that an anomaly detection system flags a user account as a bot. The system made use of lots of information about the account's activity. To a human inspecting the case, saying ``because the account left over 100 comments in the space of 60s'' may be a good explanation.

\subsection{Input Facts Disguising System Facts} \label{subsec:input-model-facts}
In cases where the user explicitly provides the input in the form of a set of facts, H2 is unlikely to select facts about the input because the user, having stated such facts themselves, presumably already believes them with high probability. Nevertheless, in such cases, an explanation containing facts about the input may seem appropriate. For example, the bank customer who provided their income and age, and asks why they were denied a loan, may be satisfied with the answer ``because your income was below such and such threshold''. However, this sort of explanation can be analysed as a form of Gricean implicature. For example, assume that the bank's policy encodes a relatively simple rule-based system for deciding whether to award loans (this could be automated or implemented by a human bank teller following a strict guide), and define:
\begin{description} \label{def:f_3_f_4}
  \item[$f_3$] The system rejects if income is $<c$.
  \item[$f_4$] Your income is $<c$.
\end{description}
The user already knows $f_4$, so $f_3$ is what a good explanation needs to convey to them. When they hear ``because $f_4$'' they can (implicitly) reason as follows: my interlocutor has presented $f_4$, which I already believe, but I also do not see how $f_4$ could explain my application being rejected. I assume they are not asserting an irrelevant fact so there must be another fact that makes this one relevant, $f_3$ is the obvious fact that would make $f_4$ relevant, so it must be that $f_3$. Thus, $f_3$ is what is actually conveyed in the sense that it is what they did not believe before and do believe afterwards. Knowing the customer will reason this way, the bank can assert $f_4$ as an explanation in place of $f_3$. Note, however, that directly asserting $f_3$ is at least as correct an explanation. An explanation using a fact about the input which the user already knew does not, therefore, contradict H2.


\section{Explanations in large language models} \label{sec:explanations-in-LLMs}
It is notoriously difficult to give explanations for LLM answers, and many neural networks in general. Our definition of a good explanation gives an account of why.

In LLMs the input is natural language, which is then converted to high dimensional vectorised representations in $\mathbb{R}^d$, undergoes many numerical operations, and is output as natural language again. At no point is the NL input decomposed into facts. Even if we restrict to NL that seeks to convey a set of facts, as opposed to the various forms of non-propositional language use, and assume humans would broadly agree on the decomposition, it is not modelled anywhere in the LLM.

Consider again the example of a bank customer applying for a loan. Suppose they write a short NL prompt expressing their age and income and ask an LLM used by the bank whether they can get a loan, and observe a single token answer: ``no''. Then they try again with an identical prompt except increasing the figure for income, and now observe the answer ``yes''. This is about as clean a case as one could hope for, and tempts us towards an explanation like ``the answer was `no' because income was below $c$'', but even here, there are several problems with this.

Firstly, it implies that ``income=$x$'' and ``age=$y$'' are the two facts in the input, but there are dozens of ways of expressing those same two facts with completely different sequences of tokens. The particular input sequence must contain other features, and so altering those features could potentially also have altered the observed output. Indeed, LLM outputs are known to be sensitive to semantically irrelevant changes to the input such as formatting \cite{sclar2023quantifying}, paraphrasing \cite{romanou2026brittlebench} or typos \cite{alahmari2025large}, so we cannot rule out these sorts of features being counterfactual explanations of the outcome. 

One could attempt to identify all the counterfactuals relating to the input by altering it in different ways and checking the answer each time. But the lack of constraints and structure on the input space again becomes a problem. What changes should be tested, and how big can a change be and still count as a single change? Maybe there is a semantically meaningless sequence of tokens that, if included anywhere, always makes the answer yes, similar to adversarial prefixes \cite{zou2023universal}. Is the absence of this sequence a counterfactual explanation? Or, assuming the question ``is Paris the capital of France?'' would have produced a ``yes'' output, should the fact that the input did not say ``is Paris the capital of France?'' be a counterfactual explanation? This would seem to have changed too many things at once about the input, reminiscent of the example from Section \ref{subsec:xor-example}, but it is hard to say in general what too much change is without knowing the set of things that can change. One option, taken e.g. by  SHAP \cite{lundberg2017unified}, is to say the individual tokens are what can change. The problem is that this restricts us to very small changes, e.g. it would likely take multiple tokens to change the figure given for income, and if we then try to allow multiple token changes, the possibilities become too numerous to check exhaustively, and also they still rule out those that involve adding new tokens. Unlike an ML system that represents the two facts explicitly, such as a decision tree, or a neural network with just two inputs (income and age), the open-endedness of LLM inputs means it is almost impossible to explore what to consider changing independently to test as possible counterfactuals. 

A second problem is that explanations will need to invoke facts about the model as well as the input. Indeed, as argued in Section \ref{subsec:input-model-facts}, an explanation like ``because income was $<c$'', is really a Gricean implication of a fact about how the model treats incomes of different levels. Trying to find counterfactuals in terms of the model weights is infeasible, there are far too many to meet any concision requirement (as per Section \ref{subsec:explanations-with-multiple-facts}). Techniques like sparse autoencoders \cite{cunningham2023sparse}, transformer circuits \cite{elhage2021mathematical} and causal intervention \cite{meng2022locating} aim to describe what is going on inside a transformer on a higher level than individual activations. However, being able to raise the user's $p(a,O)$ above some reasonable threshold would require tracing a full causal path from the input through all layers to the output, and that is far beyond current abilities for even a single input, let alone reliably across many inputs.

The only sorts of facts about LLM outputs currently usable in explanations are broad, empirically observed tendencies: why did it respond 'good question'? because most LLMs are sycophantic; why did it say its sentence contained 10 words when in fact it contained 12?, because LLMs are bad at counting. Occasionally, we can find such tendencies that qualify as facts that meet H2, but there is no guarantee we will, and for the majority of inputs we will not.

Our definition of good explanations explicates a feature of neural networks, including large language models, that is already known intuitively, namely that we cannot, as a rule, explain why they produced a given output. Our definition also implies that a requirement for an explainable model is that it can be described in terms of a set of independent facts. 

Put another way, we need a model in which we can identify $\mathcal{F}$. In the everyday examples of explanations from Section \ref{sec:proposed-definition}, humans are mostly able to identify a set of facts that are related in the appropriate way to the observation (though, as noted, further work is needed to give a theoretical account of what this appropriate relation is). In the case of explainable ML models, such as a linear model or decision tree on top of engineered features, the set $\mathcal{F}$ can largely be derived from the model's internal representation, and so our theoretical framework applies without needing further work. In the case of neural networks, the model is structured in a way that appears to mean identifying $\mathcal{F}$ will always be difficult, and we believe, though are not certain, that it will never be doable. That is, we consider it unlikely, though not impossible, that interpretability techniques reach the level where neural network inputs can be fully expressed in terms of a set of independent facts of the sorts considered by humans in explanations. This suggests that, for applications where complete explainability is essential, a model other than a neural network will be required, at least for those parts of the system whose behaviour needs to be fully explained.

\section{Conclusion}
We have presented a definition of good explanations as a form of counterfactual explanation with the additional criterion of low prior probability from the interlocutor. More work would be needed to use this definition for an analysis of explanations in general, due to the problem of identifying the set of potentially relevant facts, but for the specific case of ML explanations we can avoid this problem by restricting to facts about the model or the input. We then explored some interesting aspects of our definition, such as the case of XOR and the implication of model facts via input facts. Finally, we gave an account using this definition of the significant obstacles to explaining LLM outputs. Future work includes using this definition to guide additions/modifications of LLMs that may improve explainability.

\bibliography{bibliography}
\bibliographystyle{icml2026}

\newpage
\appendix

\end{document}